\documentclass[lettersize,journal]{IEEEtran}
\usepackage{amsmath,amsfonts}
\usepackage{algorithmic}
\usepackage{algorithm}
\usepackage{array}
\usepackage[caption=false,font=normalsize,labelfont=sf,textfont=sf]{subfig}
\usepackage{textcomp}
\usepackage{stfloats}
\usepackage{url}
\usepackage{verbatim}
\usepackage{graphicx}
\usepackage{cite}
\usepackage{orcidlink}
\usepackage{booktabs}
\usepackage{multirow}
\usepackage{makecell}
\usepackage{array}  
\usepackage{hyperref}
\usepackage{diagbox}
\usepackage{caption}
\usepackage{colortbl}
\usepackage{amssymb}
\usepackage{pifont}
\usepackage{xcolor}
\usepackage{soul}

\begin{document}

\title{Multi-Term Fourier Graph Neural Network with Sample Relationship Learning for Enhanced Remaining Useful Life Prediction}

\author{Ya Song, Laurens Bliek, Yaoxin Wu, Yingqian Zhang}

\author{
\IEEEauthorblockN{Ya Song, Laurens Bliek, Yaoxin Wu, Yingqian Zhang}

\IEEEauthorblockA{Information Systems, Department of Industrial Engineering \& Innovation Sciences, \\ Eindhoven University of Technology, Eindhoven,
The Netherlands}

\{l.bliek, y.wu2, YQZhang\}@tue.nl}

% \markboth{IEEE TRANSACTIONS ON NEURAL NETWORKS AND LEARNING SYSTEMS}%


\maketitle

\begin{abstract}
Predicting the remaining useful life (RUL) is essential for effective predictive maintenance. Spatio-Temporal Graph Neural Networks (ST-GNNs), which can model both temporal and spatial relationships by representing time series data as a sequence of graphs, have shown exceptional performance in RUL prediction. However, current ST-GNNs face several drawbacks. First, they require domain expertise or significant computational power to establish graph structures prior to deploying GNNs. Second, the models are restricted to capture temporal dependencies within a predefined fixed-size lookback window. This restriction ignores the common issue of varying time series lengths, leading the prediction model to miss short-term or long-term dependencies. Finally, conventional models often fail to capture the inherent relationships between samples generated from adjacent time windows, which are crucial for improving both the accuracy and robustness of predictions.
To address the aforementioned issues, we introduce a novel framework called Multi-Term Fourier Graph Neural Network with Sample Relationship Learning (MTFGN-SRL). Rather than treating the sample as a sequence of graphs, we consider it as a single complete graph and utilize a Fourier Graph Neural Network (FGN) to capture the spatio-temporal information in the frequency domain. We propose a multi-term learning module that utilizes multiple lookback windows to generate samples with varying terms, which are then fed into the FGN to enhance the extraction of useful information from the data. Finally, we develop a sample relationship learning module by training a heterogeneous GNN to identify inter-sample relationships, resulting in enhanced accuracy and robustness in predictions. 
Evaluations on the CMAPSS dataset demonstrate MTFGN-SRL's superior performance over state-of-the-art methods in RUL prediction.
Our codes and datasets will be made available upon publication.
\end{abstract}


\section{Introduction}
\IEEEPARstart{T}{he} widespread adoption of Cyber-Physical Systems (CPS) and the Internet of Things (IoT) allows real-time data collection and evaluation from numerous sensors integrated into machinery, enabling a more accurate and rapid assessment of equipment status. These technologies have remarkably improved the domain of predictive maintenance, providing organizations with the resources to optimize equipment management and operational effectiveness. This change aids in minimizing unexpected downtimes, decreasing maintenance expenses, and prolonging the equipment's service life.

As a proactive strategy, predictive maintenance estimates the Remaining Useful Life (RUL) of the equipment, essentially predicting the future point at which the equipment might fail or degrade in performance. It then formulates suitable maintenance plans and procedures to maintain the reliability and continuous functionality of the equipment. 
Among these activities, predicting the RUL is considered the most important, as it is both critical and highly beneficial to ensure optimal performance and prevent unexpected failures~\cite{zhou2021automatic}.
Common RUL prediction methods include those based on physical models, data-driven techniques, and hybrid approaches~\cite{ferreira2022remaining}. 

With advances in sensing technology and data analytics, data-driven approaches, particularly those based on deep learning for RUL prediction, are emerging as a vital research and application field in engineering. 
When applying deep learning methods to RUL forecasting, historical operational data and condition monitoring information are utilized. This typically includes various sources such as sensor data, operational logs, and maintenance history. Then the problem is formulated as a multivariate time series regression task, where the goal is to predict the equipment's RUL based on a combination of these data streams. 
Deep learning models rarely use raw time series data as their direct input. Instead, they start by preparing the samples with the sliding time window approach, a commonly used technique in time series analysis.
The sliding time window approach segments the time series into fixed-size overlapping windows. 
Each window contains a sequence of consecutive time points, which is then used as a single input sample for the models. As the window slides across the time series, it generates multiple samples that capture both temporal patterns and dependencies within the data. This approach ensures that the models receive structured and sequential data, facilitating better learning of temporal information.

Existing deep learning models for RUL prediction are primarily based on recurrent~\cite{da2019attention, da2020remaining,shi2021dual,wu2021degradation} and convolution neural networks~\cite{yang2019remaining,ren2020data}, and more recently on Transformer-based models that use self-attention mechanisms to dynamically assess autocorrelation~\cite{li2022domain,zhang2022dual,jiang2023new}. 
These traditional sequence models demonstrate proficiency in capturing temporal dependencies within time series data. However, they are incapable of %limited by their inability to 
considering the potential interdependencies between the various variables, limiting their effectiveness in predictions. %adversely impacting the forecasting models' effectiveness.
To mitigate this limitation, researchers have started utilizing Spatio-Temporal Graph Neural Networks (ST-GNNs)~\cite{jin2023spatio} for time series prediction. This %methodology 
involves handling data at each time step as a graph, leveraging GNNs to capture spatial information within this graph, and subsequently applying sequence models such as Transformers to derive temporal information from the resulting sequence of graph embeddings. This integrated framework significantly improves the capacity to capture complex interactions within time series data.

\begin{figure}[ht]
\centering
\includegraphics[width=0.5\textwidth]{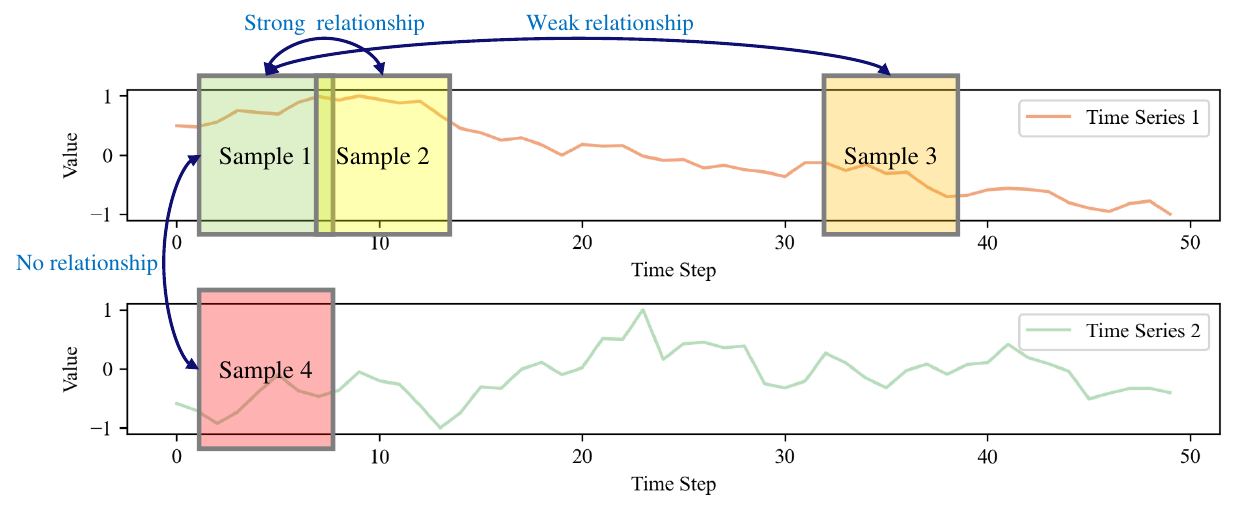 }
\captionsetup{format=plain, belowskip=5pt}
\caption{An illustrative example showing that
the different relationships among samples extracted from time series data. Temporally adjacent samples, such as Sample 1 and Sample 2, exhibit a strong relationship due to their overlapping data points and high local similarity. In contrast, samples that are farther apart, such as Sample 1 and Sample 3, demonstrate a weaker relationship, attributed to reduced temporal overlap and differing local patterns. Additionally, Sample 4 from Time Series 2 illustrates the absence of any relationship with samples from Time Series 1, as it originates from an entirely separate time series with distinct underlying dynamics. } 
\label{fig_1}
\end{figure}

Although ST-GNNs can outperform traditional sequence models in RUL prediction tasks, as shown in~\cite{kong2022spatio,wang2023comprehensive,wang2024dvgtformer}, existing ST-GNN models still exhibit four primary limitations: \\ 
(1) \textit{Requirement to learn the graph structure}. 
Unlike well-defined graph structures inherent in applications like road network traffic flow prediction, where nodes are distinctly and directly connected, the RUL prediction task faces specific challenges due to the lack of explicit interconnections between sensor signals, thereby complicating the direct application of conventional GNNs. As a result, Graph Structure Learning (GSL) methods are frequently necessary to dynamically construct an optimal graph structure that facilitates the use of GNN~\cite{wang2021spatio, chen2023convolution, wang2023comprehensive}. 
These GSL methods typically require significant computational resources. Even if an optimal graph structure is learned, conventional ST-GNNs use GNN to capture spatial information and LSTM to capture temporal information separately. This approach fails to address the potential spatio-temporal interdependencies in sensor signals. \\ (2) \textit{Fixed and short-term dependency modeling}. 
In time series prediction, the ability to model temporal dependencies over varying time horizons is crucial. Traditional approaches typically capture dependencies within a fixed-length lookback window, limiting their capacity to capture complex temporal patterns that may exist over different scales. 
As a result, these models have difficulty capturing long-term dependencies throughout the entire time series. \\ 
(3) \textit{Underutilization of test sequence information}.
Traditional RUL prediction paradigms typically focus exclusively on data from the final time window of the test sequence to make predictions, while disregarding the earlier portions of the sequence. This practice overlooks potentially valuable temporal patterns distributed across the entire test sequence. As a result, these methods fail to fully utilize the richness of the available data, limiting their ability to produce accurate and robust predictions.\\
(4) \textit{Neglect of intrinsic relationships between time series samples}. 
In machine learning-based time series forecasting, time series samples are often treated as independent observations. However, in reality, there are inherent relationships between samples from the same sequence, particularly for samples that are temporally adjacent. 
As in the example shown in Figure \ref{fig_1}, there are varying relationships among the samples extracted from the time series data.
Conventional models often fail to effectively leverage and capture these intrinsic relationships among samples, potentially resulting in less effective performance in modeling temporal dependencies.

To address these challenges, we propose an RUL prediction model called \emph{Multi-Term Fourier Graph Neural Network with Sample Relationship Learning} (MTFGN-SRL). The features and benefits of this model are outlined below. 
\begin{itemize} 
\item We introduce an innovative method for time series processing. In contrast to ST-GNNs that view a sample as a series of graphs, we regard the sample as a complete graph. Upon transforming them to the frequency domain via the Discrete Fourier Transform (DFT), we employ Fourier Graph Neural Network (FGN)~\cite{yi2024fouriergnn} to discern degradation patterns. This strategy eliminates the need for separate spatial and temporal modeling, thus facilitating the extraction of potential spatio-temporal interdependencies in sensor signal data. 
\item We propose a Multi-Term FGN (MTFGN) module to address the limited ability of traditional models in learning long-term dependencies. Specifically, the module constructs training and test graphs with varying lookback window sizes, allowing it to capture both short-term and long-term dependencies. It can optionally generate multiple predictions for ensemble averaging, thereby further enhancing predictive accuracy.
\item We develop Sample Relationship Learning (SRL) to exploit the inherent connections among time series samples with the goal of improving predictive accuracy. Within this module, we introduce a Sample Relationship Graph in which every sample acts as a node, and those from the same time series are connected. The node features are represented by embeddings learned using MTFGN. To model this graph, we design a Heterogeneous Message Passing Network, which effectively captures the complex relationships within the Sample Relationship Graph. Finally, we develop a Graph-to-Sequence Pooling method that maps the node embeddings to the corresponding labels of the time series samples, enabling accurate predictions. To the best of our knowledge, this is the first work to utilize GNNs to extract inter-sample relationships to enhance time series forecasting.
\item We test our MTFGN-SRL model on the widely used benchmark and achieve competitive performance compared to state-of-the-art traditional sequence models and ST-GNN methods. 
\end{itemize}

The rest of the paper is organized as follows. Section~\ref{sec:related} introduces the background and related works. Section~\ref{sec:method} presents the proposed framework. Section~\ref{sec:experiment} shows the experimental setting and analysis of the results. We conclude in Section~\ref{sec:conclusion}. 

\section{Background and Related Works}
\label{sec:related}
\subsection{Deep learning models for RUL prediction}
Similar to other time series forecasting applications, deep learning models have been extensively used for RUL prediction due to their ability to handle nonlinear data relationships and enable comprehensive end-to-end learning. 
The initial models focus on temporal modeling, mainly employing Long Short-Term Memory (LSTM)~\cite{shi2021dual,wu2021degradation} and Convolutional Neural Network (CNN)~\cite{yang2019remaining,ren2020data,song2021temporal} to capture temporal dependencies and recognize local patterns within time series data. Then, the Transformer~\cite{wen2022transformers} brought a significant shift in time series prediction by replacing the recurrent architecture of models like LSTMs with an entirely attention-centric mechanism. The main advancement is the self-attention mechanism, which enables the model to grasp long-range dependencies and contextual data more efficiently than recurrent structures. Researchers have also applied Transformer and elaborate variants such as Informer~\cite{zhou2021informer}, Autoformer~\cite{chen2021autoformer}, and Crossformer~\cite{zhang2023crossformer} to RUL prediction and made various improvements to the attention mechanism~\cite{li2022domain,zhang2022dual,jiang2023new}.

Recently, researchers realized that GNNs are highly effective in modeling complex relationships and dependencies between data points. They started creating graphs from the original data at each time step and utilized graph neural networks for spatial feature extraction. Key GNN architectures include Graph Convolutional Network (GCN)~\cite{wang2021spatio,wang2023comprehensive}, Graph Attention Network (GAT)~\cite{zhang2022dual,kong2022spatio}, and customized Message Passing Neural Networks (MPNN)~\cite{wang2023local}. Regarding graph construction, one study forms graphs using domain knowledge~\cite{kong2022spatio}, while another constructs adjacency matrices based on Pearson's correlation coefficients among sensors~\cite{wang2021spatio}. Recent studies tend to adopt graph structure learning methods. In \cite{chen2023convolution}, the authors describe a method of constructing the graph structure by computing the cosine similarity of the embedding vectors generated by GAT. In \cite{wang2023comprehensive}, a dynamic graph learning module is introduced to capture the evolving relationships between sensor data. 

\subsection{Multi-scale learning in time series prediction}
Multi-scale learning is widely applied in numerous fields and tasks, such as object detection and machine translation. This approach enables the model to capture features that vary in different ranges or levels of granularity and integrate local and global information~\cite{chen2021crossvit}. 
Time series data usually exhibit patterns and trends over multiple time scales, so incorporating multi-scale learning in time series analysis enables models to gain a more comprehensive understanding and capture structural information within the series~\cite{cui2016multi}. 
One common approach is employing a multi-scale convolution kernel size, simultaneously generating feature maps from receptive fields of various sizes to capture information along the temporal axis ~\cite{chen2021multi}. In~\cite{chen2023multi}, researchers applied a multi-scale pyramid network to maintain the different temporal dependencies. FTMixer~\cite{li2024ftmixer} segments the input time series into several patches of varying scales and uses a multi-scale feature fusion technique to combine feature representations from different scales. In the field of RUL prediction, the conventional multi-scale frameworks are CNNs with varying filters~\cite{ deng2022remaining, xu2022novel}.
CDSG~\cite{wang2023comprehensive} investigates the impact of varying time scales on predictions by partitioning data into patches within a lookback window, thereby establishing multiple time scales for enhanced structural understanding.
Similarly, LOGO~\cite{wang2023local} divides samples into smaller sequential patches and suggests blending global correlations with local correlations within each patch.

The multi-scale methods previously discussed aim to capture dependencies across various scales, in line with the objective of our proposed multi-term approach; nevertheless, our method adopts a unique strategy. While multi-scale methods generally retain a uniform scale for the model's input and identify relationships within this fixed dimension, our multi-term learning strategy generates multi-scale inputs through the use of different lookback window sizes. Consequently, some samples focus solely on short-term information, whereas others incorporate long-term dependencies. The following sections will elaborate on the details of this multi-term learning module.

\subsection{Sample relationship learning in time series prediction}
The conventional ST-GNNs primarily leverage GNNs to extract inter-series relationships among variables, alongside a temporal network to capture intra-series relationships. Deep Coupling Network~\cite{yi2024deep} captures the multi-order intra- and inter-series couplings of various time lags.
In the context of video object tracking, researchers have linked all inter-frame nodes across a video as a similarity graph and formulated the graph path as a Markov chain of edges~\cite{zhao2021modelling}. In addition, a component relational network was developed for time series clustering, where each entire time series is treated as a single node~\cite{li2023time}. In these studies, researchers designed models to learn the inter-series relationships among variables, the intra-series similarity relationships over consecutive time steps, or even relationships among different time series. However, the relationships among different time series samples themselves have not been considered. To the best of our knowledge, this is the first work to utilize GNNs to extract inter-sample relationships to enhance time series forecasting.

Utilizing GNNs to learn relationships between samples is rare in time series prediction but has been adopted in domains such as chemistry and biology. 
For example, researchers have developed several innovative GNNs to forecast molecular interactions~\cite{zhang2023survey}. 
GoGNN~\cite{wang2020gognn} presents the concept of a ``Graph of Graphs", comprising local graphs detailing molecules and an interaction graph depicting their dynamics. Inspired by this concept, our proposed MTFGN-SRL considers each time series sample as a complete graph and uses FGN to model spatio-temporal information. We then develop a sample relationship graph alongside a tailored GNN to capture inter-sample relationships, aiming to enhance the accuracy of time series forecasting.

\begin{figure*}[ht]
\centering
\includegraphics[width=1.0\textwidth]{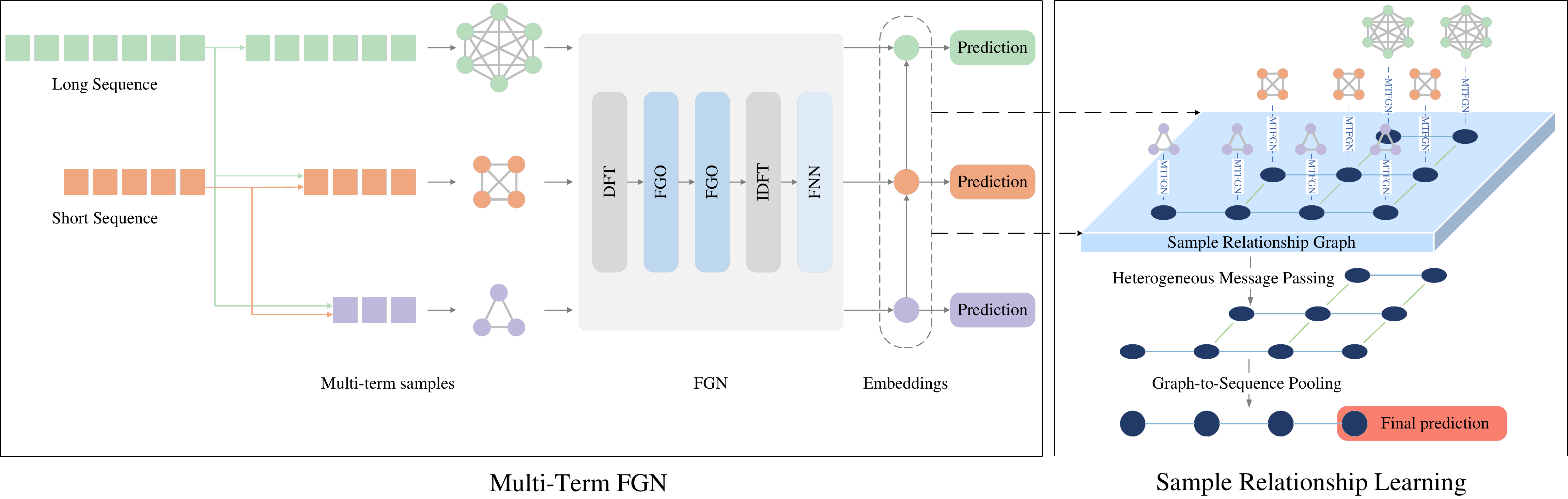}
\captionsetup{format=plain, belowskip=5pt}
\caption{Overview of MTFGN-SRL: A framework consisting of Multi-Term FGN (MTFGN) for efficient embedding generation and Sample Relationship Learning (SRL) for leveraging inter-sample relationships to boost predictive performance.}

\label{fig_2}
\end{figure*}

\section{Methodology}
\label{sec:method}
\subsection{Preliminaries and motivations}
Consider a time series dataset $\mathcal{X} = \{X^{(i)}\}_{i=1}^M$, where $X^{(i)} \in \mathbb{R}^{L_i \times N}$ represents the $i$-th sequence in the dataset, with length $L_i$ and feature dimension $N$. Specifically,
$X^{(i)} = [\mathbf{x}_1^{(i)}, \mathbf{x}_2^{(i)}, \dots, \mathbf{x}_{L_i}^{(i)}]$, 
where $\mathbf{x}_t^{(i)} \in \mathbb{R}^N$ is the feature vector at timestamp $t$ for the $i$-th time series.
Researchers commonly employ a sliding window approach to convert original time series data into samples, utilizing a lookback window of size $T$. Each generated sample consists of $T$ successive observations as input features, paired with a corresponding output label. 
The input features at timestamp $t$ for the $i$-th time series are represented as $X_t^{(i)} = \left[\mathbf{x}_{t-T+1}^{(i)}, \mathbf{x}_{t-T+2}^{(i)}, ..., \mathbf{x}_{t}^{(i)}\right]\in\mathbb{R}^{T\times N}$,  while the output label is represented as $Y_t^{(i)}$. 
The overall number of samples generated by moving the window of length $T$ across each of the $M$ time series in the dataset is represented as $\mathcal{N} = \sum_{i=1}^M (L_i - T + 1)$.
The RUL prediction task aims to forecast the label $Y_t^{(i)}$ based on input features $X_t^{(i)}$. 

When utilizing traditional sequence models to encapsulate temporal information, the prediction process can be expressed as: 
\begin{eqnarray}\label{eq1}
\hat{Y}_t^{(i)}  :=  F_{\theta_t}(X_t^{(i)})  =  F_{\theta_t}\left(\left[\mathbf{x}_{t-T+1}^{(i)}, \mathbf{x}_{t-T+2}^{(i)}, ..., \mathbf{x}_{t}^{(i)}\right]\right),
\end{eqnarray}
where $\hat{Y}_t^{(i)}$ is the predictive output corresponding to the actual value $Y_t^{(i)}$, with $F_{\theta_t}$ representing the temporal network parameterized by $\theta_t$. In the application of ST-GNN, the initial step involves the design of graphs or the employ of graph structure learning techniques to transform $\mathbf{x}_{t}^{(i)}$ into $\mathbf{g}_{t}^{(i)}$ at each time step $t$. Consequently, the RUL prediction can be articulated as:
\begin{equation}\label{eq2}
\begin{aligned}
\hat{Y}_t^{(i)} & := F_{\theta_t,\theta_g}(X_t^{(i)}) \\
& = F_{\theta_t,\theta_g}\left(\left[\mathbf{g}_{t-T+1}^{(i)}, \mathbf{g}_{t-T+2}^{(i)}, \ldots, \mathbf{g}_{t}^{(i)}\right]\right),
\end{aligned}
\end{equation}
where the forecasting function is denoted as $F_{\theta_t,\theta_g}$ parameterized by $\theta_t$ and $\theta_g$, indicating that ST-GNNs separately model temporal and spatial dependencies.

\subsection{Multi-Term FGN (MTFGN)}
We propose a Multi-Term learning module with the Fourier Graph Neural Network to tackle the deficiency of traditional models in learning long-term dependencies.  

\subsubsection{FGN for Time Series Prediction}
Recent research~\cite{yi2024fouriergnn} introduces FGN to learn unified spatio-temporal dependencies.  
FGN no longer considers input samples as a sequence of graphs; instead, it views them as one comprehensive graph. Hence, Equation \ref{eq2} can be reformulated as: 
\begin{eqnarray} 
\hat{Y}_t^{(i)}  :=  FGN_{\theta_g}(X_t^{(i)},A_t^{(i)}),
\end{eqnarray} 
where $X_t^{(i)}\in\mathbb{R}^{(T\times N) \times 1}$, $A_t^{(i)}\in \{1\}^{(T\times N) \times (T\times N)}$ represents the adjacency matrix of a complete graph, and $\theta_g$ are the parameters of the FGN. In FGN, we initially map the node features into a higher-dimensional space $d$ to obtain node embeddings $Z_t^{(i)}\in\mathbb{R}^{(T\times N) \times d}$, and perform a Discrete Fourier Transform (DFT) to transform the node embeddings into the frequency domain, resulting in $\mathcal{F}(X_t^{(i)})\in\mathbb{C}^{(\left\lfloor \frac{(T\times N)}{2} \right\rfloor + 1) \times d}$. Next, we conduct recursive multiplications between $\mathcal{F}(X_t^{(i)})$ and Fourier Graph Operators (FGOs) in the Fourier space and sum them up. Finally, we revert the node embeddings to the time domain using the Inverse Discrete Fourier Transform (IDFT), and use fully connected layers to map the embeddings to labels, as illustrated in Figure \ref{fig_2}. The detailed FGN process can be represented as follows: 
\begin{align}
\begin{split}
FGN_{\theta_g}(X_t^{(i)},A_t^{(i)}) & := \mathcal{F}^{-1} 
\left( \sum_{k=0}^{K} \sigma(\mathcal{F}(X_t^{(i)}) S_{0:k} + b_k) \right),\\
S_{0:k} & = \prod_{i=0}^{k} S_i,
\end{split}
\end{align}
where $\mathcal{F}(\cdot)$ and $\mathcal{F}^{-1}(\cdot)$ denote DFT and IDFT, respectively. $S_{k}\in\mathbb{C}^{d \times d}$ is the FGO in the $k$-th layer. $\sigma$ is the activation function, and $b_{k}\in\mathbb{C}^{d}$ are the complex-valued bias parameters. By treating time series samples as complete graphs and applying transformations in the frequency domain, FGN can effectively encode potential spatio-temporal inter-dependencies within sensor signal data while reducing noise. In addition, FGN does not require explicit learning of graph structures, making it an ideal choice for integration within our proposed module.
% add other model possible

\subsubsection{Multi-term Training Process}
When generating samples using the sliding time window method, one key parameter to set is the lookback window size, denoted as $T$. If only a single fixed window size is used, $T$ must not exceed the length of the shortest sequence in the dataset. Otherwise, some sequences would be excluded from the prediction and processing steps. This introduces the constraint $T \leq \min_{i \in \{1, 2, \dots, M\}} L_i$, where $L_i$ represents the length of the $i$-th sequence. 
However, this constraint often necessitates selecting a relatively small \(T\), which may fail to capture long-term dependencies in the data. To overcome this limitation, we propose a multi-term learning strategy. Instead of relying on a single lookback window, we utilize a set of multiple lookback windows, denoted as \[
\mathcal{T} = \{T_0, T_1, \dots, T_{C-1}\}, \quad \text{where } T_0 < T_1 < \cdots < T_{C-1},
\] to generate a diverse set of training samples. The total number of training samples is then given by:
\begin{equation}
\mathcal{N} = \sum_{c=0}^{C-1} \sum_{i=1}^M (L_i - T_c + 1).
\end{equation}

The minimum lookback window $T_0$ is set to be less than or equal to the shortest sequence length in the test dataset. This ensures that predictions can be made for all test sequences. 
Next, we gradually enlarge the window size to gather extended long-term information. By utilizing various lookback windows of differing dimensions, we can generate multi-term samples, as depicted in Figure \ref{fig_2}. This process results in $C$ groups of training samples, each corresponding to a different lookback window size, incorporating temporal dependencies at multiple scales.
During the process of creating training samples, it is possible to encounter situations where the length of a training sequence is shorter than the designated lookback window size. In such cases, we discard the entire training sequence to ensure that all training samples maintain consistent input dimensions. 
After generating the training samples, we pad the smaller samples on the left with zeros to match the length $T_{C-1}$, the largest lookback window. Then we train one FGN model in a supervised learning paradigm. This design enables the FGN model to capture diverse inherent temporal patterns in the data and enhances its ability to learn from multiple time perspectives. Once training is complete, we can already perform RUL prediction. However, instead of directly using the FGN predictions, we leverage the embeddings generated by the MTFGN as inputs to a sample relationship learning module, which is further trained for improved prediction performance.
\subsubsection{Adaptive Length-grouped Inference}
Traditional methods commonly use data from the last time window of a test sequence as the test sample, which is then fed into a trained model to obtain a single prediction result. 
In contrast, our proposed MTFGN leverages multiple lookback window sizes to generate multiple test samples from the end of each test sequence. 
These samples correspond to the same target, namely the RUL at the current point. Given that some shorter test sequences may not allow the generation of complete samples for larger time windows, our strategy is to group the test sequences according to their length and utilize applicable trained models for each group.
Figure \ref{fig_2} shows the test set divided into two groups: long and short test sequences. 
For long test sequences, multiple samples can be generated using several lookback window sizes, and these samples are then input into the trained FGN model, resulting in multiple prediction results. 
The extensive temporal information contained in long sequences allows them to leverage multiple models trained on diverse window sizes, facilitating the production of more robust and comprehensive predictions. Each prediction captures distinct temporal dependencies relevant to different time spans.
In contrast, short sequences are more limited in the amount of historical data available, which means they can only use a much smaller lookback window or only a minimum window size. Correspondingly the number of predictions that can be given is reduced.

Let the set of test sequences be denoted by ${\left\{X^{(1)}, X^{(2)}, ..., X^{(M)}\right\}}$.
For a specific test sequence $X^{(i)}$ with length $L_i$, the applicable set of lookback window sizes is determined as $T^{(i)} = \{ T_k \in T \ | \ T_k \leq L_i \}$.
For each window size $T_j \in T^{(i)}$, a test sample is generated from the last $T_j$ time steps of the sequence $X^{(i)}$. This sample is subsequently fed into the trained FGN model to generate a prediction $\hat{y}_i^{(j)}$. This results in a collection of predictions given by:
$\hat{\mathcal{Y}}_i = \{\hat{y}_i^{(1)}, \hat{y}_i^{(2)}, \ldots, \hat{y}_i^{(m)}\}$, where $m$ represents the number of valid window sizes for the sequence $X^{(i)}$. This adaptive grouping strategy optimizes the utilization of available data for each test sequence while still adhering to the constraints imposed by the sequence length. In the proposed framework, we do not directly use the RUL predictions from the MTFGN. Instead, we extract the embeddings from the penultimate linear layer of the trained FGN model during inference. These embeddings are then used as input to the Sample Relationship Learning phase.

\subsection{Sample relationship learning}

When using the sliding time window approach to generate multiple samples from time series, these samples are not independent but exhibit specific relationships. First, as the sliding window moves across the time series, there is typically some overlap between adjacent windows. This overlap implies that adjacent samples share a portion of the same data, resulting in similarities between the generated samples. Second, because of the inherent temporal correlation in time series data, the samples produced by adjacent sliding windows will exhibit similar cross-sample temporal relationships between them. Additionally, with the proposed multi-term sampling, some samples may vary in size but have the same label as they terminate at the same time point. Finally, the relationships discussed above are limited to samples generated from the same time series. In contrast, samples originating from different time series are likely to have more distant or weaker relationships.

Conventional machine learning models for time series analysis handle all time series samples as independent inputs, without considering whether these samples originate from the same time series or from different ones. These models also neglect the relationships among the adjacent samples mentioned above, and such oversight can lead to a potential decrease in prediction accuracy. 

In conclusion, it is essential to capture the relationships among time series samples. Utilizing this ubiquitous information can enhance the prediction model's ability to generate consistent predictions for adjacent samples, which in turn helps to decrease prediction errors and smooth out fluctuations over consecutive time steps.
To accomplish this goal, we first create a sample relationship graph that links all samples derived from a single time series into a graph structure. Subsequently, we introduce a novel graph neural network to comprehend the relationships among these samples within the constructed graphs.

\begin{figure}[ht]
\centering
\includegraphics[width=0.5\textwidth]{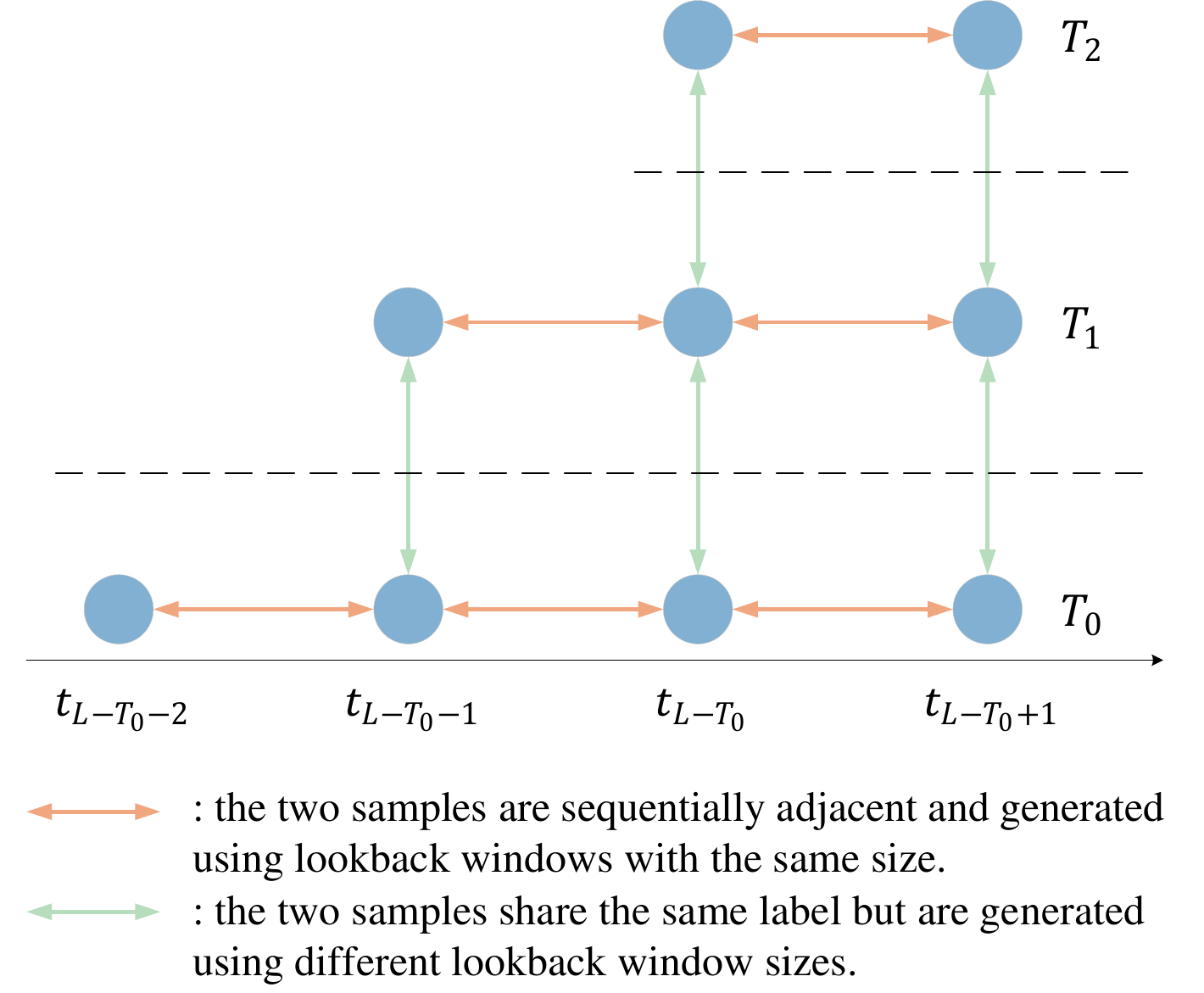}
\captionsetup{format=plain, belowskip=5pt}
\caption{An example of the proposed sample relationship graph: Representing relationships between samples generated using different lookback window sizes ($T_0$, $T_1$, $T_2$) and sequentially adjacent samples. The horizontal axis represents the timestamp, showing the temporal order of the samples.} 
\label{fig_3}
\end{figure}

\subsubsection{Sample Relationship Graph Design}

To effectively capture the inherent relationships among samples derived from a single time series, we design a Sample Relationship Graph (SRG) that integrates these samples into a structured graph representation. This graph is modeled as a heterogeneous graph to account for the distinct types of relationships among the samples, as illustrated in Figure \ref{fig_3}.

\paragraph{Node Construction} In the proposed SRG, each node represents a sample generated using the sliding time window approach. Despite the variability in sample sizes introduced by the multi-term sampling technique (which employs different lookback window sizes), the embeddings generated by the MTFGN model are uniform in size. These embeddings are used as the node features, ensuring a consistent representation across all nodes. The nodes in the SRG are homogeneous, as all nodes share the same type and are characterized solely by their corresponding sample embeddings. For the $i$-th sequence, the total number of nodes in the SRG can be calculated by summing the number of samples generated using each lookback window $T_c \in \mathcal{T}$:
\begin{equation}
N_i = \sum_{c=0}^{C-1} \left( L_i - T_c + 1 \right),
\end{equation}
where $L_i - T_c + 1$ represents the number of samples that can be generated to the sequence with a lookback window $T_c$.

\paragraph{Edge Construction} The SRG includes two distinct types of edges to represent different sample relationships, ensuring both temporal continuity and cross-scale interactions among samples. This design is motivated by the inherent characteristics of time series data, where adjacent samples often share overlapping information, and samples generated using different lookback windows can exhibit complementary relationships. 
By explicitly encoding these relationships as edges in the SRG, we create a graph structure that naturally represents the inherent dependencies in time series data, laying the foundation for efficient sample relationship learning.

\begin{itemize}
    \item \textit{Horizontal Edges:} These edges connect sequentially adjacent samples generated using lookback windows of the same size. 
    Such connections capture the temporal continuity and overlap among samples in the time series. 
    Horizontal edges allow information to flow between temporally neighboring samples, facilitating the learning of local temporal patterns and improving the consistency of predictions across time steps.
    The total number of horizontal edges for a sequence of length $L_i$ and the set of lookback windows $\mathcal{T}$ is:
    \begin{equation}
    E_H = \sum_{c=0}^{C-1} \left( L_i - T_c \right),
    \end{equation}
    % where $L_i - T_c$ represents the number of horizontal edges for the lookback window $T_c$.

    \item \textit{Vertical Edges:} These edges link samples generated using different lookback windows but terminating at the same time point. The rationale is that different lookback windows provide complementary perspectives: smaller windows capture short-term dependencies, while larger windows highlight long-term trends. Vertical edges connect these samples, enabling the model to aggregate multi-term information and leverage the complementary nature of short-term and long-term dependencies effectively. 
    The total number of vertical edges is:
    \begin{equation}
    E_V = \sum_{c=1}^{C-1} \left( L_i - T_c + 1 \right),
    \end{equation}
\end{itemize}

Combining the horizontal and vertical edges, the total number of edges in the SRG for the $i$-th sequence is:
\begin{equation}
E_i = E_H + E_V = \sum_{c=0}^{C-1} \left( L_i - T_c \right) + \sum_{c=1}^{C-1} \left( L_i - T_c + 1 \right).
\end{equation}

The SRG's edge structure is crafted to illustrate the variety of connections found in time series data. 
Given the heterogeneity of these edges, the SRG is designed as a heterogeneous graph, where the two edge types explicitly encode different relational semantics. Notably, the edges in the SRG do not have additional feature values; their primary role is to delineate the structural relationships among the samples. 
By organizing the samples into this well-structured SRG, the graph representation enables effective learning of relationships within the same time series while isolating samples from different series. This design supports the propagation and aggregation of information among related samples, laying a solid foundation for sample relationship learning. 

\subsubsection{Heterogeneous Message Passing Network}

To effectively learn the sample relationships encoded in the SRG, we design a Heterogeneous Message Passing Network (HMPN) tailored to the structure of the graph. This network utilizes distinct message-passing operations to process the horizontal and vertical edges, effectively capturing the temporal and cross-scale relationships represented in the SRG. By integrating information from both types of edges, The HMPN operates in three main stages:

\paragraph{Horizontal-based Convolution} Horizontal edges connect sequentially adjacent samples generated using the same lookback window size. These edges capture the temporal continuity inherent in the data. To process this information, we perform a horizontal-based convolution, which aggregates information from a node's horizontal neighbors. The updated representation for node $v$ is computed as:
\begin{equation}
\mathbf{h}_v^{(\text{H})} = \phi_{\text{H}}\left(\mathbf{h}_v, \Psi_{\text{H}}\left(\left\{\psi_{\text{H}}\left(\mathbf{h}_u\right) \mid u \in \mathcal{N}_{\text{H}}(v) \right\}\right)\right),
\end{equation}
where $\mathcal{N}_{\text{H}}(v)$ represents the set of horizontal neighbors of node $v$, while $\mathbf{h}_v$ and $\mathbf{h}_u$ denote the feature vectors of node $v$ and its neighboring node $u$, respectively. The functions $\phi_{\text{H}}$ and $\psi_{\text{H}}$ are transformations specific to horizontal edges. The function $\Psi_{\text{H}}$ is a permutation-invariant aggregation function that combines the transformed features from the neighboring nodes. 

\paragraph{Vertical-based Convolution} Vertical edges link samples generated using different lookback windows but terminating at the same time point. These edges reflect the complementary nature of short-term and long-term dependencies. To aggregate information from vertical neighbors, we perform a vertical-based convolution, updating the representation of node $v$ as:
\begin{equation}
\mathbf{h}_v^{(\text{V})} = \phi_{\text{V}}\left(\mathbf{h}_v, \Psi_{\text{V}}\left(\left\{\psi_{\text{V}}\left(\mathbf{h}_u\right) \mid u \in \mathcal{N}_{\text{V}}(v) \right\}\right)\right),
\end{equation}
where $\mathcal{N}_{\text{V}}(v)$ represents the set of vertical neighbors of node $v$, and $\mathbf{h}_v$ and $\mathbf{h}_u$ denote the feature vectors of node $v$ and its neighboring node $u$. The functions $\phi_{\text{V}}$, $\psi_{\text{V}}$, and $\Psi_{\text{V}}$ are the vertical counterparts of $\phi_{\text{H}}$, $\psi_{\text{H}}$, and $\Psi_{\text{H}}$, respectively, and are designed to capture relationships specific to vertical edges.
This operation enables the model to effectively aggregate multi-term information, integrating perspectives from both short-term and long-term dependencies.

\paragraph{Max Aggregation} After performing horizontal-based and vertical-based convolutions, the node representations need to be combined to fully exploit the heterogeneous structure of the SRG. We use a max aggregation strategy, which ensures that the most salient features from both types of convolutions are retained. The final node representation is computed as:
\begin{equation}
\mathbf{h}_v = \max \left( \mathbf{h}_v^{(\text{H})}, \mathbf{h}_v^{(\text{V})} \right),
\end{equation}
where the max operation is applied element-wise to the feature vectors. This strategy effectively merges the contributions of both temporal continuity (horizontal edges) and cross-scale relationships (vertical edges), resulting in a robust representation for each node.

\begin{figure}[ht]
\centering
\includegraphics[width=0.5\textwidth]{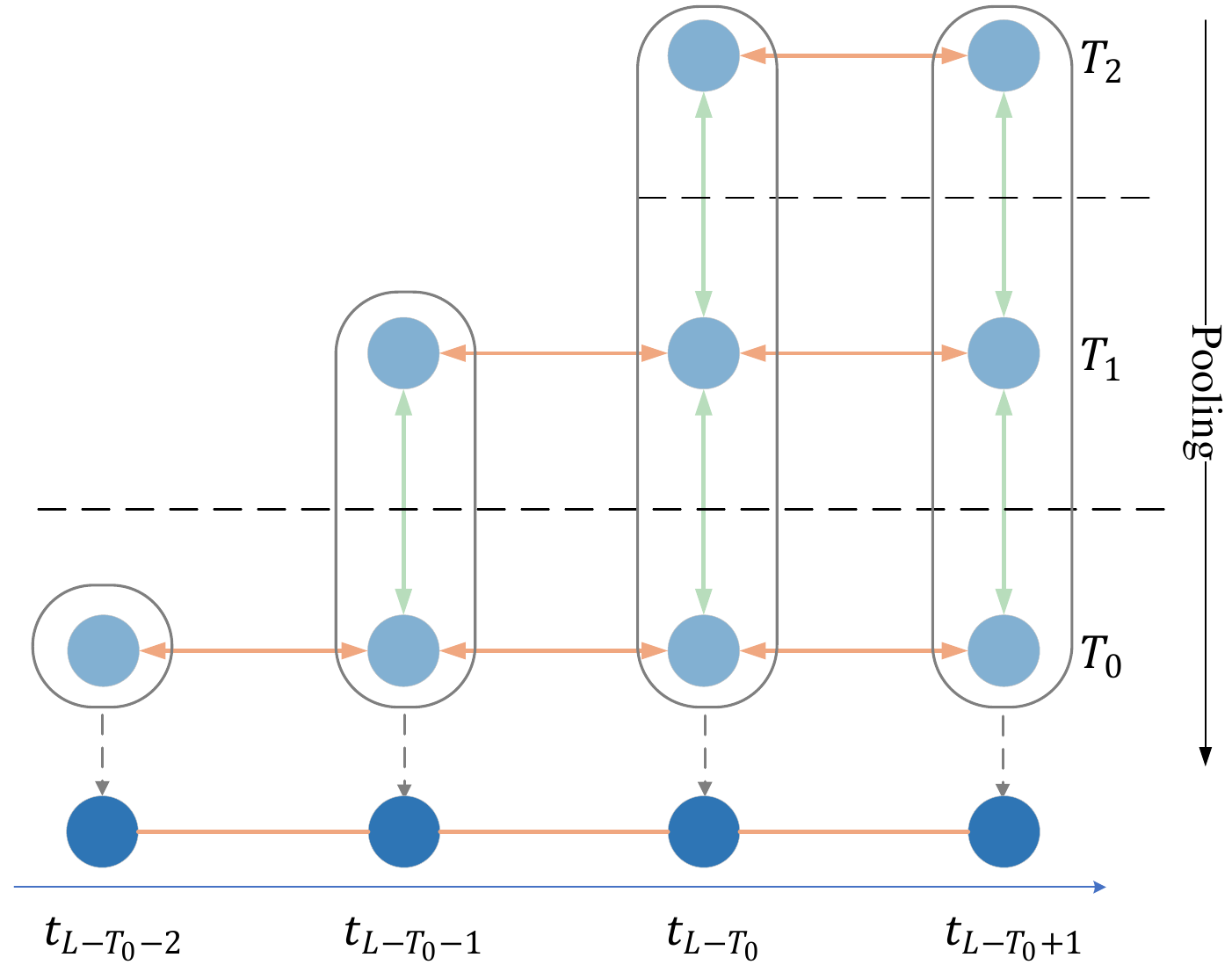 }
\captionsetup{format=plain, belowskip=5pt}
\caption{Graph-to-Sequence Pooling: Aggregating node representations via average pooling along the vertical axis to generate predictions for each time step.} 
\label{fig_4}
\end{figure}

\subsubsection{Graph-to-Sequence Pooling}
Upon acquiring node embeddings via HMPN, we develop a Graph-to-Sequence Pooling method to convert these node embeddings into a sequence of representations corresponding to each time step. We achieve this by performing average pooling along the vertical axis of the graph, aggregating the embeddings of nodes associated with the same time step, as shown in Figure \ref{fig_4}. Formally, let $\mathcal{N}_i^t$ denote the set of nodes associated with time step $t$ in sequence $i$, and $\mathbf{h}_v$ represent the embedding of node $v$ obtained from HMPN. The pooled representation for time step $t$, denoted as $\mathbf{p}_t$, is computed as:
\begin{equation}
\mathbf{p}_t = \frac{1}{|\mathcal{N}_i^t|} \sum_{v \in \mathcal{N}_i^t} \mathbf{h}_v,
\end{equation}
where $|\mathcal{N}_i^t|$ is the number of nodes corresponding to time step $t$ in sequence $i$.

This pooling operation integrates multi-term information from nodes generated with different lookback windows, capturing the temporal and relational context encoded in the graph. By averaging node embeddings, it also reduces noise and variability, resulting in more robust representations for each time step.
The resulting pooled representations, $\{\mathbf{p}_t\}_{t=0}^{L_i - T_0}$, form a sequence of embeddings, where each $\mathbf{p}_t$ corresponds to the aggregated information for time step $t$. These embeddings are subsequently used as inputs for generating predictions for each time step. 

To map the sequence of embeddings to predictions, we primarily use a Multi-Layer Perceptron (MLP). However, since the pooled representations form a sequence that retains temporal alignment, we can also utilize LSTM to further capture the temporal dependencies among the samples. Specifically, the LSTM processes the sequence of pooled representations $\{\mathbf{p}_t\}_{t=0}^{L_i - T_0}$ as:
\begin{equation}
\mathbf{h}_t = \text{LSTM}(\mathbf{p}_t, \mathbf{h}_{t-1}),
\end{equation}
where $\mathbf{h}_t$ is the hidden state at time step $t$, and $\mathbf{h}_{t-1}$ is the hidden state from the previous time step. The final predictions for each time step can then be derived by applying a fully connected layer to the LSTM output:
\begin{equation}
\hat{\mathbf{y}}_t = \mathbf{W}_{\text{out}} \mathbf{h}_t + \mathbf{b}_{\text{out}},
\end{equation}
where $\mathbf{W}_{\text{out}}$ and $\mathbf{b}_{\text{out}}$ are learnable parameters of the output layer.

This design allows the framework to flexibly leverage either MLP or LSTM for the prediction task. While MLP provides a straightforward mapping from pooled representations to predictions, LSTM can effectively model temporal dependencies among the time steps, further enhancing the model’s capability to capture sequential patterns and improve prediction accuracy.

\section{Experiments}
\label{sec:experiment}
In this section, we thoroughly evaluate the proposed framework on a benchmark dataset.

\subsection{Dataset description}
% Data Description and Preprocessing evaluation metrics
The Commercial Modular Aero-Propulsion System Simulation (CMAPSS) dataset is a well-known public dataset widely used in the field of Remaining Useful Life (RUL) prediction~\cite{xia2020ensemble}. It comprises four distinct subsets.
Each subset is divided into a training set and a test set. The training set includes multiple instances of turbofan engine condition monitoring data, ranging from normal operation to total failure. In contrast, the condition monitoring data in the test set stops before reaching complete failure. The objective is to predict the RUL of the engines in the test set.
Table \ref{tab_1} provides the details of the characteristics of each subset. Of the four datasets (FD001-FD004), the engines in FD001 and FD003 operated under a single operational condition while those in FD002 and FD004 operated under six different operational conditions, making predictions more complex. Furthermore, FD001 and FD002 engines have one fault mode, specifically the High-Pressure Compressor (HPC) failure, whereas FD003 and FD004 each have two fault modes.
Table \ref{tab_1} also shows the minimum and maximum sequence lengths within the dataset, highlighting the significant variability in sequence lengths across different engines.

The training data capture engine operations up until the point of failure, resulting in relatively long signal records. In addition, differences in initial engine states and failure processes lead to varying sequence lengths for each engine.
Existing models typically use a fixed lookback window to generate samples~\cite{kong2022spatio,wang2021spatio,chen2023convolution}. However, the size of this lookback window cannot exceed the shortest sequence length of the test engines; otherwise, the model cannot provide predictions for all test engines.
This constraint on the lookback window size is unsuitable for test engines with relatively long sensor data sequences, potentially limiting the model's ability to learn long-term dependencies.

\begin{table}[ht]
\renewcommand\arraystretch{1.2}
\caption{Description of CMAPSS turbofan engine dataset.}\label{tab_1}
\centering
\setlength{\tabcolsep}{1mm}{
\begin{tabular}{cccccccc} \hline
Subset & \makecell{Operation\\Conditions} & \makecell{Fault\\Mode} &  \makecell{Training\\sequence} &  \makecell{Test\\sequence} & \makecell{Maximum\\length} & \makecell{Minimum\\length}\\ \hline

FD001  & 1 & HPC & 100  & 100  & 362 & 31 \\
FD002  & 6 & HPC & 260  & 259  & 378 & 21\\
FD003  & 1 & HPC+Fan & 100 & 100  & 525 & 38\\
FD004  & 6 & HPC+Fan & 249 & 248  & 543 & 19\\ \hline
\end{tabular}}
\end{table}

\subsection{Implementation settings}
\begin{figure*}[ht]
\centering
\includegraphics[width=1.0\textwidth]{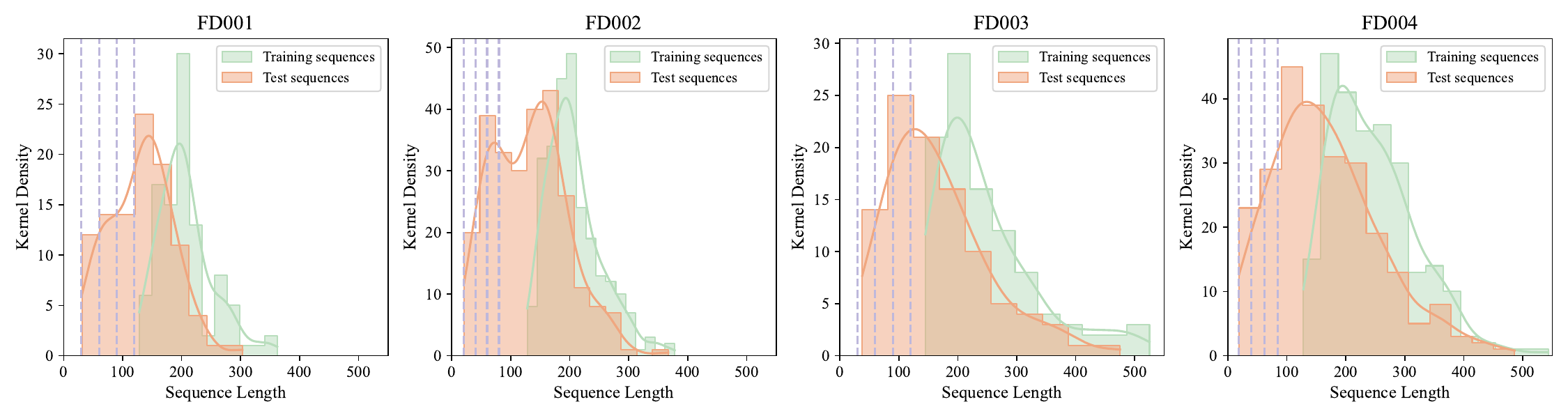}
\captionsetup{format=plain, belowskip=5pt}
\caption{The training and test sequence length distribution of four subsets in the CMAPSS dataset, multiple dashed lines parallel to the y-axis represent the various sizes of the lookback windows we employed. These dashed lines partition the test sequences into multiple subgroups.} 
\label{fig_5}
\end{figure*}

We ensure consistency in data preprocessing settings as in ~\cite{kong2022spatio, wang2021spatio} to maintain fair comparisons with existing models. Initially, we normalize the 14 effective features chosen from the original set of 24 features. We then use a piecewise function to adjust the training and test labels, capping them at 125 to avoid RUL overestimation. 
Next, we implement our proposed multi-term learning approach by using multiple lookback windows to create various samples. As shown in Table \ref{tab_2}, we use different lookback window sizes for the four subsets. Unlike existing methods that employ a single window size, which is constrained by the shortest sequence length in the subset and results in short samples that cannot capture long-term dependencies, we use a range of lookback window sizes. We set the initial window size smaller than the shortest sequence length to ensure the model's applicability, then gradually increase the window size to produce longer samples, enabling the model to capture potential long-term dependencies. 
The lookback time windows we use and their comparison with the time series length are shown in Figure \ref{fig_5}. 
We divide the training and testing sequences into subgroups using the defined multiple lookback windows. Training and testing samples are generated adaptively according to the sequence length within each subgroup. With the relatively longer sequences on the right side of each sub-figure, all the lookback windows on the left side of the sequences can be used, whereas the reverse is not valid.

\begin{table}[ht]
\renewcommand\arraystretch{1.2}
\caption{Comparison of multiple time windows used in the proposed framework against the single time window setting in existing works.}\label{tab_2}
\centering
\setlength{\tabcolsep}{1mm}{
\begin{tabular}{cccc} \hline
Subsets  & Min length & Single window size & Window sizes in MTFGN-SRL\\ \hline
FD001  & 31 & 30 & 30/60/90/120 \\
FD002  & 21 & 20 & 20/40/60/80\\
FD003  & 38 & 30 & 30/60/90/120\\
FD004  & 19 & 15 & 18/40/62/84\\ \hline
\end{tabular}}
\end{table}

Given that FD002 and FD004 data operate under six distinct conditions, using FGN directly for frequency-domain learning could result in a low signal-to-noise ratio. We identified six features highly correlated with the labels [``s7", ``s9", ``s11", ``s12", and ``s13"] ~\cite{huang2023unsupervised} and proceeded to cluster the data with k-means method and normalize it under each cluster. Next, FGN was employed to learn from these input samples. 
We configured the number of FGO layers to three, which suffices for the RUL prediction task.
In SRL, we employ a single-layer graph convolution. For FD002 and FD004, whose sequences are longer, we incorporate an LSTM-based mapping to better capture their extended temporal dependencies. All implementation details and hyperparameter settings are provided in our publicly available code repository.
Two evaluation metrics were utilized: the Root Mean Square Error (RMSE) and a Score function ~\cite{kong2022spatio}, defined by the following equation:
\begin{eqnarray}
\text{Score}(v, \hat{v}_i) = 
\begin{cases} 
\sum_{i=1}^{M}(e^{-\frac{\hat{v}_i - v_i}{13}} - 1) & \text{if } \hat{v}_i < v_i; \\
\sum_{i=1}^{M}(e^{\frac{\hat{v}_i - v_i}{10}} - 1) & \text{if } \hat{v}_i \geq v_i,
\end{cases}
\end{eqnarray}
where $v_i$ and $\hat{v}_i$ represent the true and predicted RUL values, respectively. The asymmetric Score function assigns a higher penalty for overestimating RUL, as overestimated RULs entail more severe consequences. Similar to RMSE, a lower score function value indicates better prediction performance.

\subsection{Comparisons with state-of-the-art}
This section compares our method with the most advanced RUL prediction techniques available~\cite{wang2023local}. We primarily focus on ST-GNNs due to their strong performance on this task. The benchmarked approaches can be broadly categorized into two groups. The first group consists of 5 sequence models, with Transformer-based models being the most prominent. The second group comprises 12 ST-GNN based models, which leverage spatial information to achieve superior predictive performance, generally surpassing sequence models. 
Our approach, MTFGN-SRL, differs significantly from existing methods in several key ways. First, it conducts learning in the frequency domain rather than the time domain, transforming samples into a graph instead of a sequence of graphs. Second, it adopts a multi-term learning strategy to enhance the model's ability to capture long-term dependencies within samples. Lastly, it introduces a sample relationship learning procedure, further improving performance by modeling relationships across samples.

Table~\ref{tab_3} compares the RMSE and Score values of MTFGN-SRL with other advanced sequence models and ST-GNNs on the CMAPSS dataset. The results for the benchmarked models are sourced directly from their respective original papers for consistency and accuracy.
The results clearly demonstrate that MTFGN-SRL achieves state-of-the-art performance across all four subsets (FD001–FD004) of the dataset. Specifically, MTFGN-SRL achieves the lowest average RMSE and Score, outperforming all baseline methods. 
Compared to the second-best method, LOGO, MTFGN-SRL reduces the average RMSE and Score by 11.6\% and 21.6\%, respectively. This significant improvement highlights the effectiveness of incorporating sample relationship learning to capture critical dependencies across samples and enhance predictive robustness.
These results validate the effectiveness of MTFGN-SRL in addressing the challenges of RUL prediction, particularly in its ability to capture both short-term and long-term dependencies, as well as sample relationships. These attributes make MTFGN-SRL a state-of-the-art approach for predictive maintenance tasks.

\begin{table*}[htb]
\renewcommand\arraystretch{1.2}
\caption{Comparison of RMSE and Score values for MTFGN-SRL, advanced sequence models, and ST-GNN models on the CMAPSS dataset  (bold: best; underline: runner-up).}\label{tab_3}
\centering
\setlength{\tabcolsep}{3mm}{
\begin{tabular}{c|cccccccc|cc} 
\hline
\multirow{2}{*}{Models} & \multicolumn{2}{c}{FD001} & \multicolumn{2}{c}{FD002} & \multicolumn{2}{c}{FD003} & \multicolumn{2}{c}{FD004} & \multicolumn{2}{c}{Average}\\ 
 & RMSE & Score & RMSE & Score & RMSE & Score & RMSE & Score & RMSE & Score\\
\hline
% from https://www.sciencedirect.com/science/article/pii/S0951832024002369#bib0036
DA-Transformer~\cite{liu2022aircraft}  & 12.25 & 198 & 17.08 & 1575 & 13.39 & 290 & 19.86 & 1741 & 15.65 & 951.00 \\
BiGRU-TSAM~\cite{zhang2022prediction} & 12.56 & 213 & 18.94 & 2264 & 12.45 & 233 & 20.47 & 3610 & 16.11 & 1580.00 \\
MSIDSN~\cite{zhao2023multi} & 11.74 & 206 &	18.26 & 2047 &	12.04 & 196 &	22.48 & 2911 & 16.13 & 1340.00 \\
EAPN~\cite{zhang2023predicting} & 12.11 & 245 & 15.68 & 1127 & 12.52 & 267 & 18.12 & 2051  & 14.61 & 922.50 \\
Crossformer~\cite{wang2023local} & 12.11 & 216 & 14.16 & 837 & 12.32 & 260 & 14.81 & \underline{956} & 13.35 & 567.25 \\
\hline

HAGCN~\cite{li2021hierarchical} & 11.93 & 222 & 15.05 & 1144 & 11.53 & 240 & 15.74 & 1219 & 13.56 & 706.25 \\

STGCN~\cite{wang2021spatio} & 14.55 & 402 & 14.58 & 943 & 13.06 & 394 & 14.60 & 1065 & 14.20 & 701.00 \\

% 从DVGTformer文章来
STFA~\cite{kong2022spatio} & 11.35 & 194 & 19.17 & 2493 & 11.64 & 225 & 21.41 & 2760 & 15.89 & 1418.00 \\
DAST~\cite{zhang2022dual} & 11.43 & 203 & 15.25 & 925 & 11.32 & \textbf{155} & 18.36 & 1491 & 14.09 & 693.50 \\
GGCN~\cite{wang2022gated} & 11.82	& 187 & 17.24 & 1494 &	12.21 & 245 & 17.36 & 1372 & 14.66 & 824.50 \\
ConvGAT~\cite{chen2023convolution} & 11.34 & 197 & 14.12 & \underline{772} & 10.97 & 235 & 15.51 & 1231 & \underline{12.99} & 608.75 \\

CDSG~\cite{wang2023comprehensive} & \underline{11.26} & 188 & 18.13 & 1740 & 12.03 & 218 & 19.73 & 2332 & 15.29 & 1119.50 \\

DCFA~\cite{gao2023dual} & 11.74 & 190 & 16.81 & 1076 & \underline{10.71} & 198 & 17.77 & 1571 & 14.26 & 758.75 \\
LOGO~\cite{wang2023local}& 12.13 & 226 & \underline{13.54} & 832 & 12.18 & 261 & \underline{14.29} & \underline{944} & 13.04 & \underline{565.75} \\
NSD-TGTN~\cite{gao2024nonlinear} & 12.13 & 226 & 15.87 & 1477 & 12.01 & 220 & 16.64 & 1493 & 14.16 & 854.00 \\

% TKGIN (2023) & 10.21 & x.xx & 11.56 & x.xx & 10.17 & x.xx & 12.43 & x.xx \\
DVGTformer~\cite{wang2024dvgtformer} & 11.33 & \underline{180} & 14.28 & 797 & 11.89 & 255 & 15.50 & 1108 & 13.25 & 585.00 \\

THGNN~\cite{wen2024temporal} & 13.15 & 285 & 13.84 & 806 & 12.61 & 255 & 14.65 & 1166 & 13.56 & 628.00 \\
\hline

MTFGN-SRL& \textbf{10.39} & \textbf{160} & \textbf{11.39} & \textbf{562} & \textbf{10.19} & \underline{180} & \textbf{14.10} & \textbf{874}  & \textbf{11.52} & \textbf{444.00} \\

\hline
\end{tabular}}
\end{table*}

\subsection{Ablation study}

\begin{table*}[htb]
\renewcommand\arraystretch{1.2}
\caption{Ablation study on the CMAPSS dataset with average performance across all datasets.}\label{tab_4}
\centering
\setlength{\tabcolsep}{2.0mm}{
\begin{tabular}{c|ccc|cccccccc|cc} 
\hline
\multirow{2}{*}{Variants} & \multirow{2}{*}{FGN} & \multirow{2}{*}{\makecell{MT}} & \multirow{2}{*}{\makecell{SRL}} & \multicolumn{2}{c}{FD001} & \multicolumn{2}{c}{FD002} & \multicolumn{2}{c}{FD003} & \multicolumn{2}{c}{FD004} & \multicolumn{2}{c}{Average} \\
 & & & & RMSE & Score & RMSE & Score & RMSE & Score & RMSE & Score & RMSE & Score \\
\hline
CNN  & \ding{55} & \ding{55} & \ding{55} & 12.91 & 236 & 15.88 & 1002 & 12.05 & 185 & 18.27 & 1847 & 14.78 & 817.50 \\
MTCNN  & \ding{55} & \ding{51} & \ding{55} & 13.09 & 285 & 12.39 & 628 & 11.49 & 223 & 14.81 & 1068 & 12.95 & 551.00 \\
MTCNN-SRL  & \ding{55} & \ding{51} & \ding{51} & 12.91 & 276 & 12.03 & 578 & \underline{11.15} & 196 & \underline{14.12} & \textbf{807} & 12.55 & \underline{464.25} \\
FGN  & \ding{51} & \ding{55} & \ding{55} & 11.91 & 190 & 15.30 & 1123 & 11.93 & \textbf{176} & 18.38 & 1842 & 14.38 & 832.75 \\
MTFGN  & \ding{51} & \ding{51} & \ding{55} & \underline{10.74} & \underline{188} & \underline{11.88} & \underline{562} & 11.46 & 254 & 14.85 & 1003 & \underline{12.23} & 501.75 \\
MTFGN-SRL  & \ding{51} & \ding{51} & \ding{51} & \textbf{10.39} & \textbf{160} & \textbf{11.39} & \textbf{562} & \textbf{10.19} & \underline{180} & \textbf{14.10} & \underline{874} & \textbf{11.52} & \textbf{444.00} \\
\hline
\end{tabular}}
\caption*{\footnotesize MT: Multi-Term learning module, SRL: Sample Relationship Learning module.}
\end{table*}

To rigorously evaluate our framework components, we conducted ablation studies on the CMAPSS dataset using two base architectures (CNN and FGN), while systematically analyzing the impacts of Multi-Term (MT) and Sample Relationship Learning (SRL). Our experimental design features three key configurations: 1) Base architectures, 2) Standalone MT implementation that generates multi-term predictions through ensemble averaging, and 3) Combined MT+SRL integration. This structure enables clear attribution of performance improvements to either multi-term temporal modeling (via MT) or inter-sample relationship learning (via SRL) when working with different backbone architectures. Table~\ref{tab_4} presents the performance metrics (RMSE and Score) across all subsets for different model variants.

First, the findings illustrate that FGN serves as a robust baseline, persistently yielding lower prediction errors compared to the traditional CNN model over most of the subsets of the dataset. This confirms its elevated ability to handle intricate spatio-temporal patterns via spectral decomposition.
Incorporating the MT module results in significant improvements for both CNN and FGN architectures, with FGN implementations demonstrating increased responsiveness. This consistent enhancement across various architectures confirms the efficacy of our MT module. Significantly, the improvement greatly diminishes prediction errors in long-sequence scenarios (FD002/FD004), validating the module's efficacy in capturing long-term temporal dependencies.
The SRL module also demonstrates strong performance across different base architectures, and its impact amplifies when combined with FGN architecture. The most striking improvement emerges in FD003, where MTFGN-SRL achieves 11.08\% lower RMSE than the standalone MTFGN. Interestingly, MTCNN-SRL also shows competitive results on FD004 and FD002. This is particularly significant as it demonstrates that even with a simple CNN architecture, our proposed modules (MT and SRL) can elevate the model's performance to surpass many sophisticated STGNN architectures.

In terms of computational efficiency, FGN requires 4.99 seconds per epoch on an NVIDIA V100 GPU, which exceeds the CNN baseline of 1.35 seconds, yet requires only 270k trainable weights compared to CNN's 511k. 
Integrating the Multi-Term learning module slightly increases training time to 5.48 seconds per epoch and reduces parameters to 24k while substantially boosting accuracy by capturing multi-term dependencies.
The decrease in model parameters is primarily attributed to the enhanced generalization capability enabled by multi-term sample inputs.
The SRL module introduces additional computational costs for graph processing, requiring 2.25 seconds per epoch but further enhances performance by leveraging sample relationships. Overall, our framework achieves its superior performance through more efficient architectural design and better feature learning rather than simply increasing model complexity.
The ablation study demonstrates that each component of our framework contributes meaningfully to the final performance. Furthermore, the impressive results of MTCNN-SRL highlight that our modules are architecture-agnostic and can effectively enhance even simple baseline architectures to achieve competitive performance.

\begin{figure}[ht]
\centering
\includegraphics[width=0.5\textwidth]{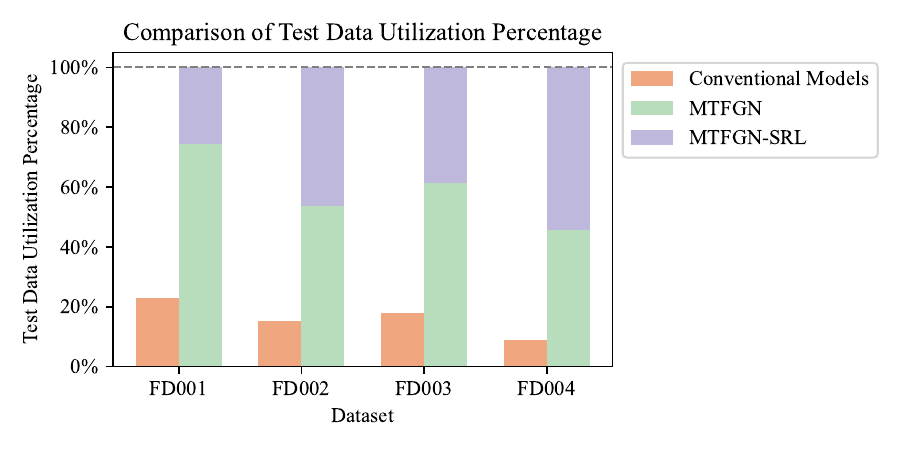}
\captionsetup{format=plain, belowskip=5pt}
\caption{Comparison of test data utilization percentage across the CMAPSS dataset.} 
\label{fig_6}
\end{figure}

\begin{figure}[ht]
\centering
\includegraphics[width=0.5\textwidth]{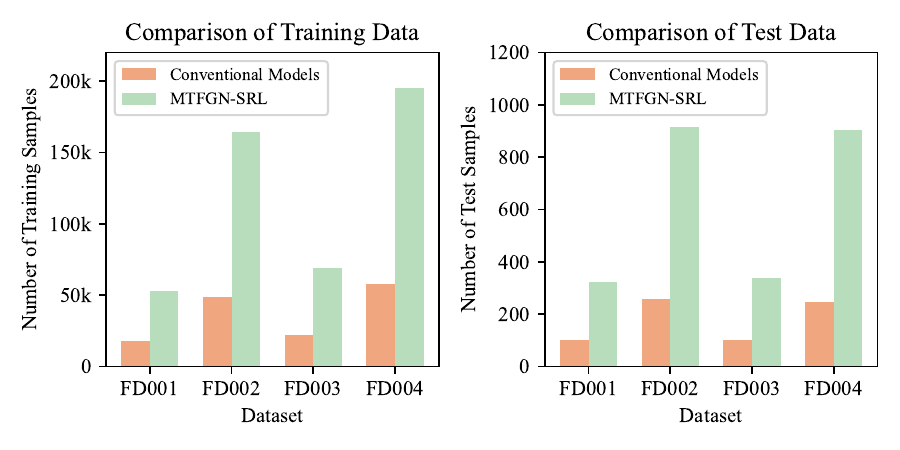}
\captionsetup{format=plain, belowskip=5pt}
\caption{Comparison of the number of training and test samples across the CMAPSS dataset.} 
\label{fig_7}
\end{figure}

\subsection{Data utilization analysis}
Our framework significantly improves data utilization efficiency by leveraging a greater portion of the available information in the time series data.
Figure~\ref{fig_6} illustrates the comparison of test data utilization rates within the CMAPSS dataset. Conventional approaches demonstrate significant underuse of available test data, primarily due to their reliance on single fixed-length sliding windows with limited temporal scope to generate test samples. 
MTFGN significantly improves data utilization by integrating multiple lookback window sizes, allowing the model to capture temporal dependencies at different time scales within test sequences. More crucially, the SRL component achieves 100\% data utilization across all subsets through its ability to model comprehensive relationships among samples throughout the sequence.
Although this effectively leverages the complete temporal context, it is important to note that historical data indirectly influence RUL predictions, through a graph structure that encodes sample relationships and enables the learning of more informative sample representations, ultimately leading to enhanced prediction accuracy.

Figure~\ref{fig_7} compares the number of training and test samples generated by conventional models and our proposed approach across the CMAPSS dataset. The results highlight that our framework generates significantly more samples with different scales, both for training and testing, compared to conventional methods. This is particularly evident in larger subsets, such as FD002 and FD004, where our model generates a much richer set of samples. The increased sample generation stems from the use of multi-term lookback windows and the ability of SRL to model overlapping and related samples effectively. The resulting enriched sample space not only increases training data diversity but also ensures comprehensive coverage of underlying temporal patterns, culminating in superior predictive performance.

\subsection{Robust prediction analysis}

\begin{figure*}[ht]
\centering
\includegraphics[width=1.0\textwidth]{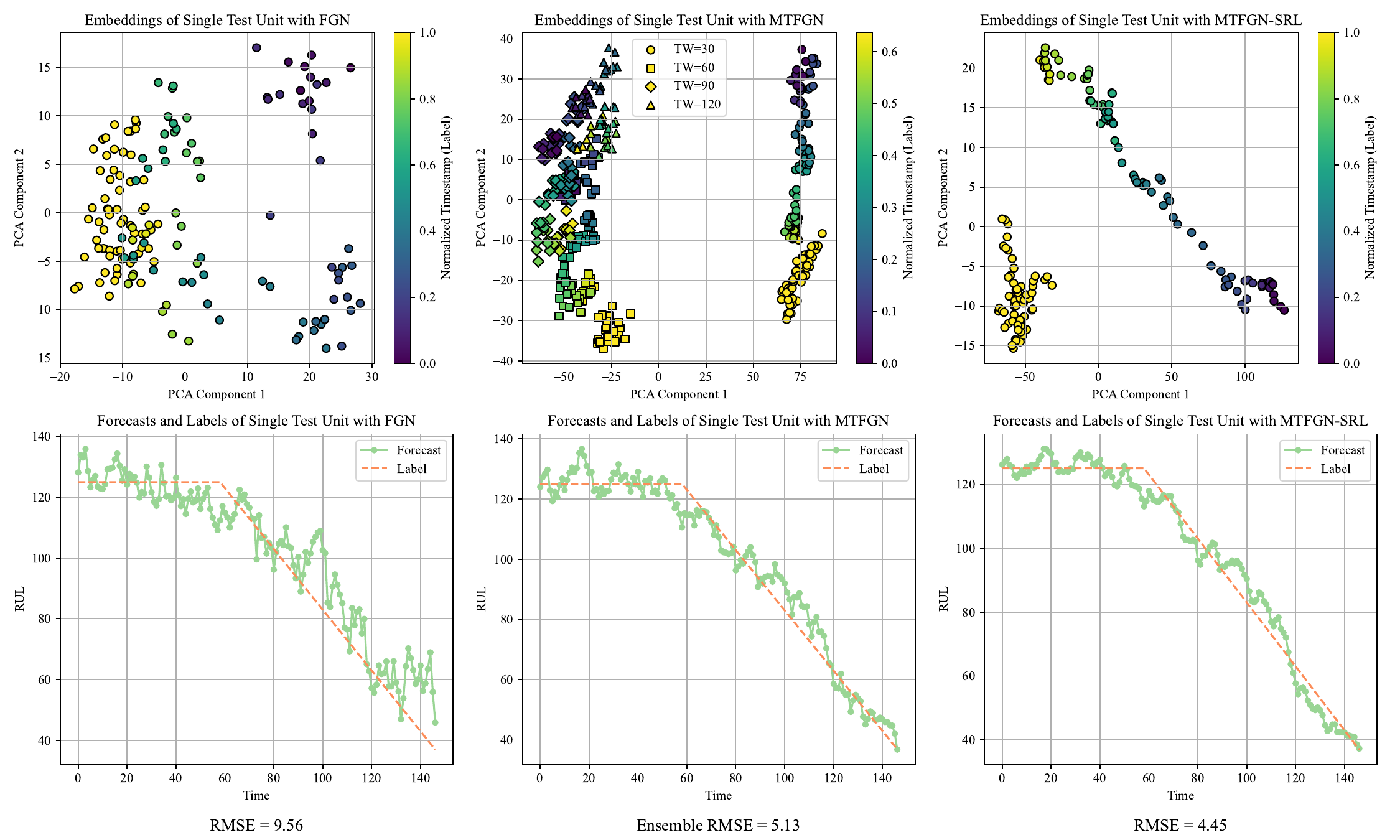}
\captionsetup{format=plain, belowskip=5pt}
\caption{Feature embeddings and prediction performance comparison across models (FGN, MTFGN, and MTFGN-SRL) on FD001 test unit (ID = 58). The top row shows Principal Component Analysis (PCA) reduced feature embeddings with colors indicating normalized RUL labels. The bottom row compares predicted RUL values to ground truth, highlighting that MTFGN and MTFGN-SRL reduce fluctuations and provide more stable predictions than FGN.}
\label{fig_8}
\end{figure*}

% new
Our framework is designed to reduce the risk of isolated predictions and enhance the coherence of learned representations. Through comprehensive visualization and analysis presented in Figure~\ref{fig_8}, we demonstrate the progressive improvements achieved by incorporating multi-term learning and sample relationship learning components, examining both feature embeddings and prediction results across three model variants: FGN, MTFGN, and MTFGN-SRL.

Figure~\ref{fig_8} visualizes feature embeddings in the top row, which reveals distinct characteristics across the three models.
The baseline FGN model, which processes samples independently, generates embeddings that exhibit uniform scatter without recognizable temporal structure, indicating its inability to capture sequential patterns in the data. The embeddings generated by MTFGN exhibit clearer temporal trends due to the use of multi-term learning. Notably, embeddings corresponding to \( \text{TW}=30 \) are distinctly separated from others, as this window size does not require zero-padding, preserving the original structure of the data. However, the most significant advancement is demonstrated by MTFGN-SRL, which produces highly structured embeddings that clearly delineate the temporal degradation trajectory, particularly as the RUL decreases. This enhanced coherence stems from SRL's explicit modeling of inter-sample relationships, resulting in representations that faithfully capture the temporal evolution of the system state.

The bottom row of Figure~\ref{fig_8} presents the prediction results, which further demonstrates the superior performance of our proposed approach.
While FGN's predictions display significant instability and deviation from the true RUL trajectory, MTFGN achieves moderate improvement through its multi-term module, though some inconsistencies persist. MTFGN-SRL demonstrates remarkable stability and accuracy in its predictions, maintaining consistent alignment with the ground truth trajectory. This enhanced performance can be attributed to SRL's ability to leverage information across related samples, ensuring predictions remain coherent with the underlying temporal structure.

\section{Conclusion}
\label{sec:conclusion}
In this paper, we proposed a novel framework called the Multi-Term Fourier Graph Neural Network with Sample Relationship Learning (MTFGN-SRL), to address critical challenges in RUL prediction. Unlike traditional ST-GNNs that require predefined graph structures and are limited by fixed-size lookback windows, MTFGN-SRL leverages an FGN to model the spatio-temporal dependencies in the frequency domain. This approach eliminates the need for explicit graph structure learning.
To further improve the model's capability to capture long-term dependencies, we introduced a Multi-Term FGN (MTFGN) module. This module constructs training and test graphs with varying lookback window sizes, enabling FGN to capture both short-term and long-term dependencies.
Moreover, we developed a heterogeneous graph neural network to construct a Sample Relationship Graph (SRG) and learn the inherent relationships between samples generated from sliding time windows. This sample relationship learning framework ensures that the model captures both temporal continuity and cross-window relationships, leading to smoother and more robust predictions.
The proposed MTFGN-SRL was evaluated on the CMAPSS benchmark dataset and achieved superior performance compared to state-of-the-art methods, demonstrating its effectiveness in improving the accuracy and robustness of RUL prediction. By integrating frequency-domain modeling, multi-term sampling, and sample relationship learning, this framework provides a powerful solution for predictive maintenance tasks and sets a new standard for RUL prediction models.

\bibliographystyle{IEEEtran}
\bibliography{IEEEabrv,mybibfile}

\vfill

\end{document}